\documentclass[letterpaper, 10 pt, conference]{ieeeconf}  

\IEEEoverridecommandlockouts                              

\usepackage{textcomp}
\usepackage{gensymb}
\usepackage{url}
\usepackage{siunitx}
\usepackage{units}
\usepackage{colortbl}
\usepackage{color}
\usepackage{multicol} 
\usepackage{caption}
\usepackage{subcaption}
\usepackage{xargs}  
\usepackage{pgfplots}
\usepackage{floatrow}   
\usepackage{graphicx,xcolor}

\usepackage{amsmath} 
\usepackage[noadjust]{cite}
\usepackage{hyperref}

\title{\LARGE \bf
Flexible Trinocular: Non-rigid Multi-Camera-IMU Dense Reconstruction for UAV Navigation and Mapping
}

\author{Timo Hinzmann, Cesar Cadena, Juan Nieto, and Roland Siegwart
\thanks{*This work was supported by the \emph{Federal office armasuisse Science and Technology} under project n\textdegree 050-45.}
\thanks{All authors are with the ETH, the Swiss Federal Institute of Technology Zurich,
Autonomous Systems Lab (www.asl.ethz.ch),
Leonhardstrasse 21, LEE,
CH-8092 Zurich, Switzerland.
{\tt \{\underline{firstname.lastname}}\}@mavt.ethz.ch.}
}

\begin{document}
\begin{minipage}{\textwidth}
\copyright 2019 IEEE. Personal use of this material is permitted. Permission from IEEE must be obtained for all
other uses, in any current or future media, including reprinting/republishing this material for advertising
or promotional purposes, creating new collective works, for resale or redistribution to servers or lists, or
reuse of any copyrighted component of this work in other works.\\

Please cite this paper as:\\%
\begin{verbatim}
@InProceedings{Hinzmann2019_trinocular,
  Title                    = {Flexible Trinocular: Non-rigid Multi-Camera-IMU Dense
                              Reconstruction for UAV Navigation and Mapping},
  Author                   = {Hinzmann, Timo and Cadena, Cesar and Nieto, Juan and 
                              Siegwart, Roland},
  Booktitle                = {2019 {IEEE/RSJ} International Conference on
                              Intelligent Robots and Systems ({IROS}},
  Year                     = {2019},
  Organization             = {IEEE}
}

\end{verbatim}

\end{minipage}

\maketitle
\thispagestyle{empty}
\pagestyle{empty}

\begin{abstract}
In this paper, we propose a visual-inertial framework able to efficiently estimate the camera poses of a non-rigid trinocular baseline for long-range depth estimation on-board a fast moving aerial platform.
The estimation of the time-varying baseline is based on relative inertial measurements, a photometric relative pose optimizer, and a probabilistic wing model fused in an efficient Extended Kalman Filter (EKF) formulation.
The estimated depth measurements can be integrated into a geo-referenced global map to render a reconstruction of the environment useful for local replanning algorithms.
Based on extensive real-world experiments we describe the challenges and solutions for obtaining the probabilistic wing model, reliable relative inertial measurements, and vision-based relative pose updates and demonstrate the computational efficiency and robustness of the overall system under challenging conditions.
%
\end{abstract}

\section{INTRODUCTION}
\begin{figure}[t]
\centering
\includegraphics[width=\linewidth]{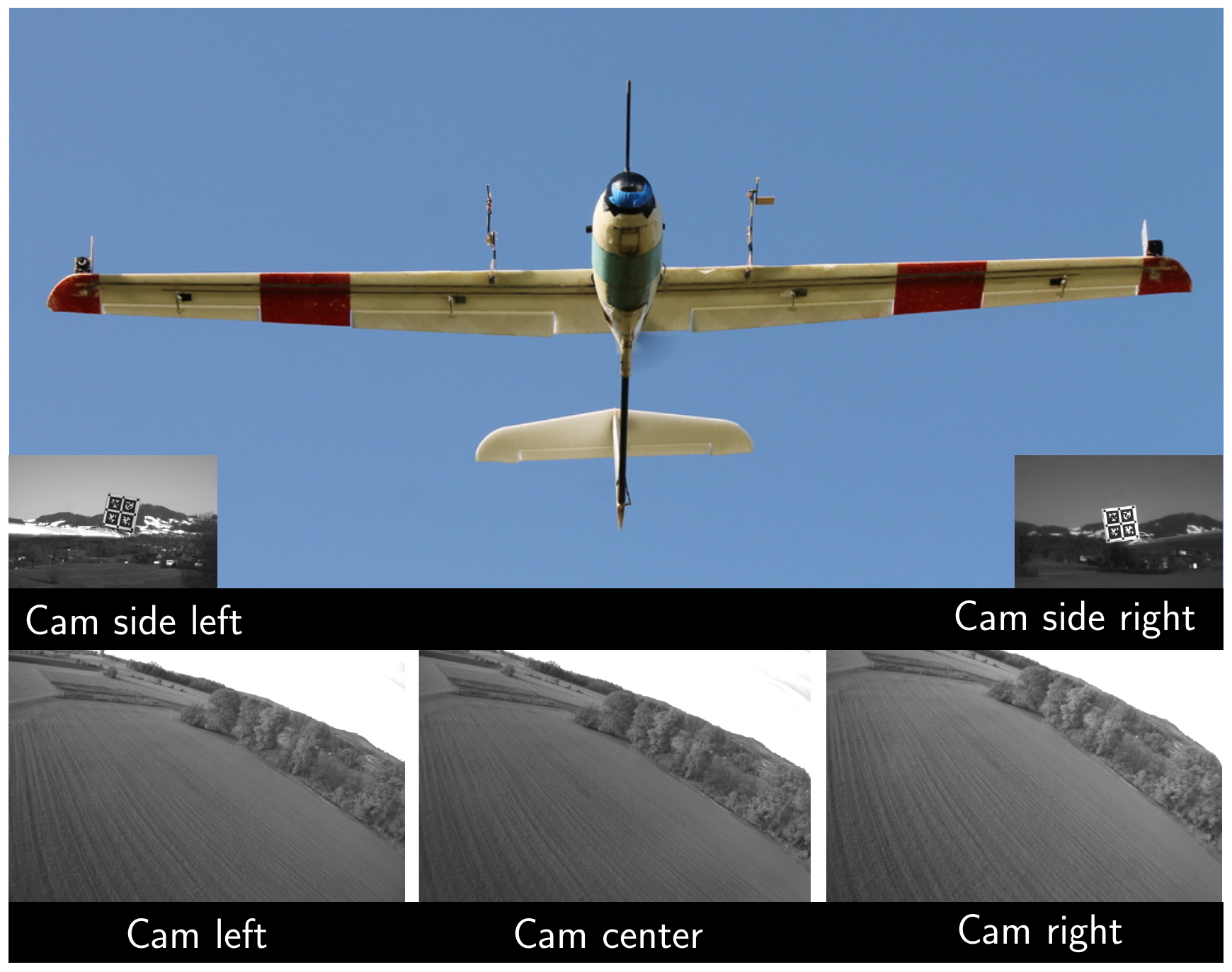}
\caption{Fixed-wing UAV platform \emph{Techpod} equipped with the proposed trinocular visual-inertial sensor setup. The side cameras and April tags are used for the identification of the wing model.}
\label{fig:teaser}
\end{figure}

For environmental awareness and autonomous mission execution in previously unknown terrain unmanned aerial platforms depend on high-quality depth measurements and ideally have access to a globally consistent map.
For ground robots and rotary-wing Unmanned Aerial Vehicles (UAVs), small baseline rigid stereo rigs or depth cameras have shown to be sufficient to enable autonomous operation.
However, for small fixed-wing UAVs flying at \SI{15}{\meter/\second}, like the platform shown in Fig. \ref{fig:teaser}, off-the-shelf sensor solutions fail for various reasons: 
Monocular depth from motion (high depth uncertainty due to epipole), small baseline rigid stereo (limited range and depth uncertainty), laser (limited range, heavy, and expensive), depth cameras (limited range), radar (not yet miniaturized).
In this paper, we present our approach to increase the environmental awareness of fixed-wing UAVs to efficiently estimate the time-varying relative pose between multiple cameras while only relying on inexpensive visual-inertial sensors.
%
In contrast to \cite{Hinzmann2018}, in this paper we propose a trinocular setup where the center camera is rigidly mounted with respect to the autopilot located inside the fuselage to simplify the generation of a geo-referenced and globally consistent map.
Additionally, the trinocular setup allows to take advantage of both the full baseline for long-range depth estimates (left to right wing tip camera; approx. $\SI{2.4}{\metre}$ for our platform shown in Fig. \ref{fig:teaser}) and half the baseline (e.g. left to center camera) for obstacles nearby or for landing procedures.
\begin{figure}[b!]
\centering
\includegraphics[width=0.49\linewidth]{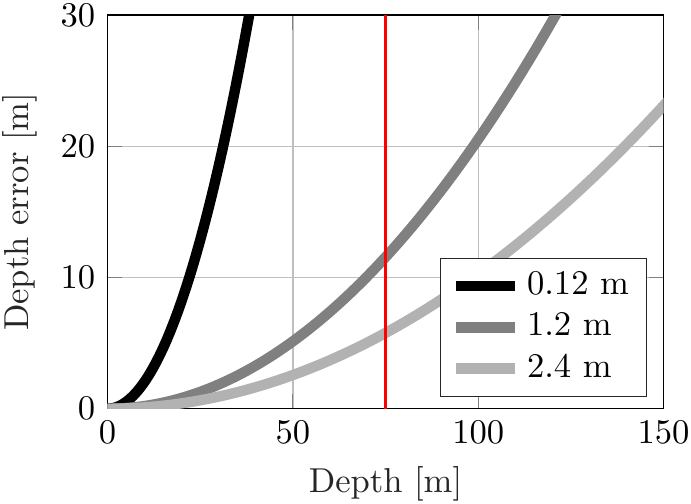} 
\hfill
\includegraphics[width=0.49\linewidth]{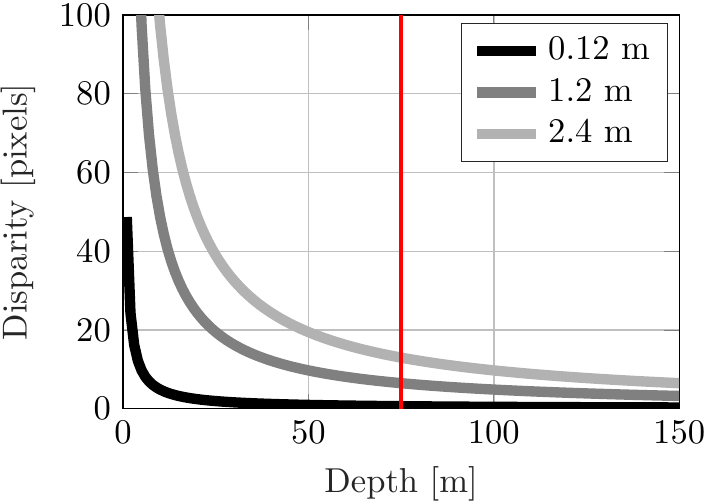}
\caption{Theoretical depth error and disparity for different stereo baselines in meters. The red vertical line marks the distance that our UAV platform can cover within \SI{5}{\s}. Further parameters: focal length of \SI{2.8}{\milli\metre}, pixel size of \SI{6}{\micro\metre}, assumed resolvable disparity of one pixel.}
\label{fig:theory}
\end{figure}
We estimate the non-rigid time-varying baseline between the center and left/right camera with Extended Kalman Filters (EKFs) that fuse relative visual-inertial measurements with a calibrated wing model that further constrains the relative baseline transformation. 
The depth maps resulting from the multiple stereo views can then be integrated into a geo-referenced map.
While the approach presented in \cite{Hinzmann2018} is interesting for reactive navigation and control of a UAV, the framework proposed in this paper is designed for the use of model predictive controllers and local replanning algorithms able to operate on a $\unit[3]{D}$ or $\unit[2.5]{D}$ map.
\begin{figure*}[htb]
  \includegraphics[width=\textwidth]{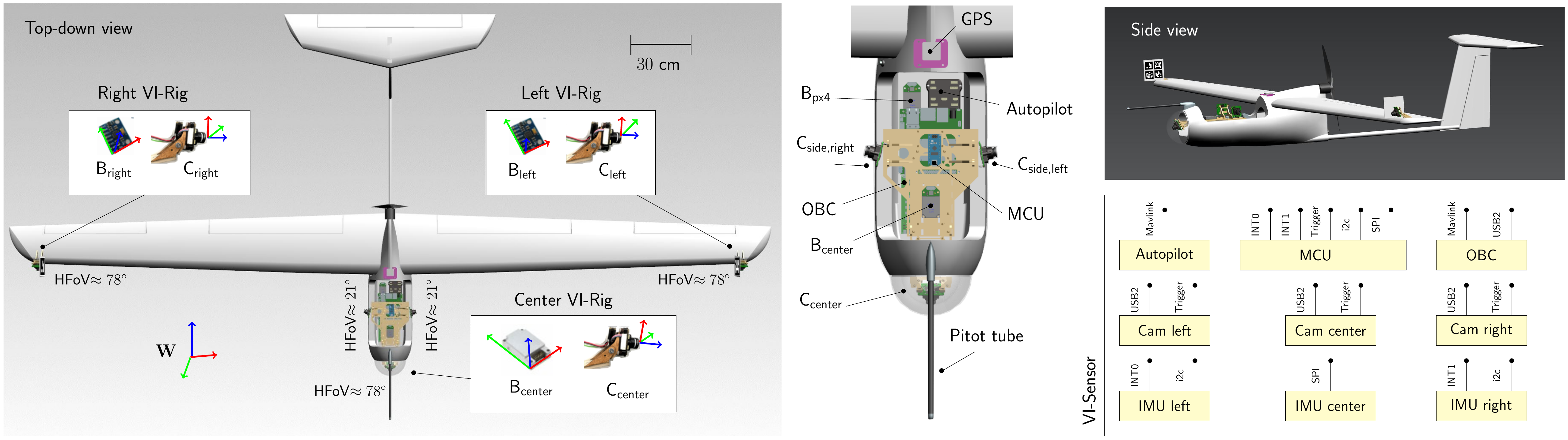}
  \caption{Overview showing the UAV platform, sensors, and coordinate frames. An EKF estimates the relative pose between left, right, and center camera. The side cameras are used to identify a wing model in form of a Gaussian prior. The horizontal field of view (HFoV) is denoted for each camera. The schematic on the right bottom depicts the developed VI-sensor that time-synchronizes the sensors on a hardware level.}
  \label{fig:overview}
\end{figure*}
\section{RELATED WORK}
%
%
Several state-of-the-art monocular visual-inertial odometry (VIO) and simultaneous localization and mapping (SLAM) frameworks have been adopted to stereo or multi-camera operation \cite{Bloesch2015, leutenegger2015keyframe, Mur-Artal2017, Houben2016, Sun2018}.
Other multi-camera systems are specifically designed for their individual application such as for increased environmental awareness to navigate in cluttered environments using a multi-copter \cite{Gohl, Miiller2018} or for obstacle detection and avoidance on-board an agile delta wing \cite{Barry2015}. 
All of these approaches have in common that they assume a fixed and precisely calibrated transformation between the individual cameras.
However, for a fixed-wing UAV as shown in Fig.~\ref{fig:teaser}, a \emph{rigid} stereo system can, due to aerodynamic effects and payload constraints, only be mounted inside the fuselage, limiting the effective baseline in this case to approximately \SI{12}{\centi\metre}.
Fig. \ref{fig:theory} presents the theoretical depth uncertainty and expected disparity for such a rigid small-baseline stereo setup and motivates the necessity for a setup with a wider baseline.

On the other hand, approaches have been developed that are able to cope with different degrees of deviations from the calibrated baseline transformation:  Warren et al. \cite{warren2013online} continuously estimate thermally induced slow changes in the baseline.
In \cite{warren2016long}, a down-looking stereo pair with a wide-baseline (\SI{0.7}{\metre}) is employed and the relative transform between the cameras, together with the poses of the stereo rig itself is estimated in an offline bundle adjustment problem.
Lanier et al. \cite{lanier2011modal, lanier2010, Short2009a} use a set of accelerometers distributed along the wing to estimate vibrational disturbances without any vision update.
For airborne applications, Yang et al. fuse the measurements from a master and slave IMU in an EKF to estimate the relative states, a process referred to as transfer alignment \cite{Yang2019}.
Achtelik et al. \cite{achtelik2011collaborative} proposed an efficient EKF formulation that fuses relative IMU and vision measurements for stereo vision with two quadrocopters that have an overlapping field of view.
Recently, this EKF formulation has been used in \cite{Hinzmann2018, Teixeira2018, Karrer2018} for various applications.
In \cite{Hinzmann2018}, the initial concept of \emph{Flexible Stereo} was introduced, further constraining the EKF formulation with a probabilistic wing model.
The approach allows obstacle avoidance with a reactive controller as the depth map is computed with respect to the left (or right) wing tip camera.

In contrast to \cite{Hinzmann2018}, this paper suggests to use a trinocular setup to enable the generation of a geo-referenced and globally consistent map. Furthermore, we suggest a vision update that utilizes photometric feature tracking in combination  with relative inertial data to obtain an improved initial position.  
Whilst \cite{Hinzmann2018} is only tested in simulation, this publication demonstrates the real-world application.
\section{CONTRIBUTIONS AND ASSUMPTIONS}
This paper incorporates recent advances in relative visual-inertial state estimation \cite{Achtelik2011, Hinzmann2018, Forster2017} and contains the following contributions:
\begin{itemize}
\item A procedure to obtain a probabilistic wing model that considers the aerodynamic forces acting on the wing during the flight. 
\item Implementation and validation of an efficient framework to estimate the time-varying relative transformation between multiple stereo setups.
\item Real world experiments, including platform design and hardware considerations.
\end{itemize}
%
To focus on the description of the above-mentioned contributions we assume that the center camera is aligned with the map throughout the paper.
%
%
For flying in GPS-denied environments this can be realized for instance with \cite{Hinzmann2016} or \cite{Forster2017}.
As the state estimator implemented on our autopilot already provides pose priors in real-time we instead employ the approach suggested in \cite{hinzmann2016robust} but augment it with IMU edges.
In summary, the proposed framework assumes the following input: 1) features with corresponding depth estimates tracked in the center camera for relative camera-camera alignment, 2)  pose estimates of the center camera in the map frame for depth map registration.
\section{THE APPROACH}
In Sec. \ref{secA} we present our approach to align the wing camera's pose to the center camera.
All steps in this section follow the paradigm of staying close to the optimal relative pose to increase the efficiency of the photometric image alignment and relative pose estimation.
Sec. \ref{secB} shortly summarizes the generation of the depth maps while Sec. \ref{secC} describes our procedure to obtain a probabilistic wing model.
\subsection{Inertial-photometric alignment of wing to center camera}
\label{secA}
The pose of a wing camera is aligned with the center camera using an EKF in relative formulation \cite{Achtelik2011}, efficiently fusing relative IMU measurements, photometric visual cues, and a probabilistic wing model \cite{Hinzmann2018}.
The state vector is defined as follows
\begin{align}
 \mathbf{x} &= \left[ \mbox{$\bar q_c^j$}, \; {\mathbf{\omega}_c^c}, \; {\mathbf{\omega}_j^j}, \; {\mathbf{p}_c^j}, \; {_w \mathbf{v}_c^j}, \; {_w \mathbf{a}_c^c}, \; {_w \mathbf{a}_j^j} , \; \mathbf{b_a}^j_c, \; \mathbf{b_\omega}^j_c \right]^\top
\end{align}
with $j\in\{\text{left}, \text{right}\}$. 
In comparison to \cite{Achtelik2011, Hinzmann2018}, the state  is augmented to compensate for the IMU bias.
Formulating the relative pose alignment in a relative EKF is efficient and modular, allowing to easily add additional visual-inertial sensor rigs to, e.g., further increase the overall field of view.

Based on the last iteration's estimate of $\mathbf{T}^{C_{j,k-1}}_{C_{c,k-1}}$, the relative transformation is propagated to the time stamp of the current stereo frame using the IMU measurements of the center and wing camera (as outlined in \cite{Achtelik2011, Hinzmann2018}), resulting in an initial prior for $\mathbf{T}^{C_{j,k}}_{C_{c,k}}$ at time step $k$.
This relative pose prior is then used to project features observed in the center image into the wing camera via
\begin{align}
d &= \| \mathbf{p}^{W}_{\text{lm}} - \mathbf{p}^{W}_{C_{c,k}}\|_2 \nonumber  \\
\mathbf{u}_j &= \Pi(\mathbf{T}^{C_{j,k}}_{C_{c,k}} \cdot \mathbf{C}_c \cdot  d)
\label{eq:depth}
\end{align}
where $\mathbf{p}^{W}_{\text{lm}}$ is the estimated landmark position, $d$ is the observed sparse depth estimate corresponding to this landmark,  $\Pi(\cdot)$ is the projection operator, and $\mathbf{C}_c$ is the bearing vector from the center camera to the landmark.
Patches around the expected set of feature positions $\mathbf{u}_j$ are extracted from the wing camera image and the relative transformation aligning the wing to the center camera is further refined using sparse image alignment.
The photometric error is minimized as proposed in \cite{Forster2014, Forster2017}, using a constrained Gauss-Newton solver with pyramidal layers:
\begin{align}
&\mathbf{T}^{C_{j, k}}_{C_{c, k}} 
= \arg\min_{\mathbf{T}^{C_{j, k}}_{C_{c, k}}}
\frac{1}{2}\sum_{i}\|\delta\mathbf{I}(\mathbf{T}^{C_{j,k}}_{C_{c,k}} ,\mathbf{u}_{j,i})\|^2_{\Sigma_I} \nonumber \\
&+ \frac{1}{2}\|\mathbf{p}^{C_{j,k}}_{C_{c,k}} - \bar{\mathbf{p}}^{C_j}_{C_c}\|_{\Sigma_\mathbf{p}}
+ \frac{1}{2}\|\log (\bar{\mathbf{R}}^{C_j}_{C_c}{}^\top \mathbf{R}^{C_{j,k}}_{C_{c,k}})^\vee\|_{\Sigma_\mathbf{R}}.
\label{eq:sparse_img_align}
\end{align}
%
%
For the derivation and notation details we refer to \cite{Forster2014, Forster2017}.
As motion constraint we choose the transformation prior $\bar{\mathbf{T}}^{C_j}_{C_c}=(\bar{\mathbf{R}}^{C_j}_{C_c}, \bar{\mathbf{p}}^{C_j}_{C_c})$ obtained from the wing model.
Alternatively, the current mean and covariance from the EKF could be used. 
%
%
%
While the translation computed in the vision update presented in \cite{Achtelik2011, Hinzmann2018} is only up to scale, the approach described in this publication computes a full-scale translation.
The relative pose from the vision step is then fused in the EKF with the Gaussian wing model as in \cite{Hinzmann2018}, for the case that the vision step fails, and the framework continues with the next iteration. 

\paragraph{Discussion}
In our experiments, above approach has shown to be a robust way to transfer and track features from one to another camera under challenging conditions (e.g. different illumination or exposure time) as the feature locations are constrained by the relative transformation and are not tracked individually.
On the other hand, tracking each feature patch individually, for instance with a Lucas-Kanade (LK) tracker, was not robust and resulted in many outlier matches.
Note that this LK tracker implementation showed a good performance for establishing time-sequential frame-to-frame  correspondences for the \emph{same} camera.
%
%

%
%
%
\subsection{Dense reconstruction}
\label{secB}
The optimized camera poses are used to rectify the images using the algorithm proposed in \cite{Fusiello2000} with subsequent stereo block-matching (BM).
Based on the three cameras, the following combinations of depth maps can be computed:
$\mathbf{d}^c_{c,l}$, $\mathbf{d}^c_{c,r}$, and $\mathbf{d}^l_{l,r}$ with their different characteristics visualized in Fig. \ref{fig:theory}.
For instance $\mathbf{d}^c_{c,l}$ denotes the depth map computed from the left and center image, expressed in the center camera.
%

%
%
%
\subsection{Obtaining a wing model}
\label{secC}
\begin{figure}[htb]
  \begin{subfigure}[b]{0.49\linewidth}
    \includegraphics[width=\textwidth]{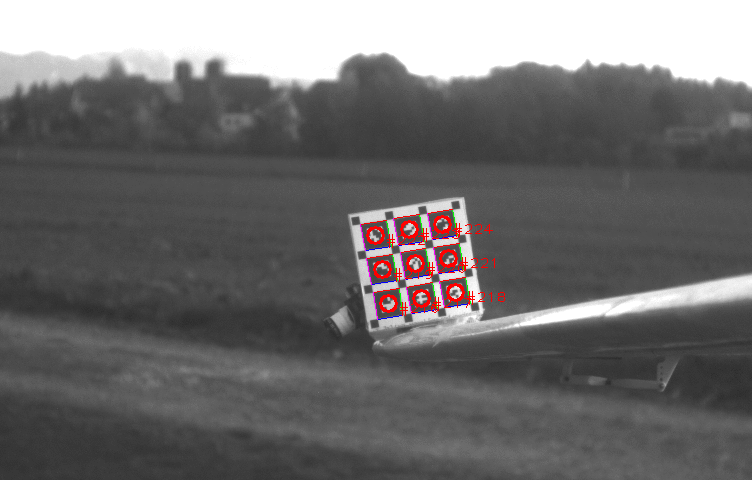}
    \caption{On-ground}
  \end{subfigure}
  \hfill
  \begin{subfigure}[b]{0.49\linewidth}
    \includegraphics[width=\textwidth]{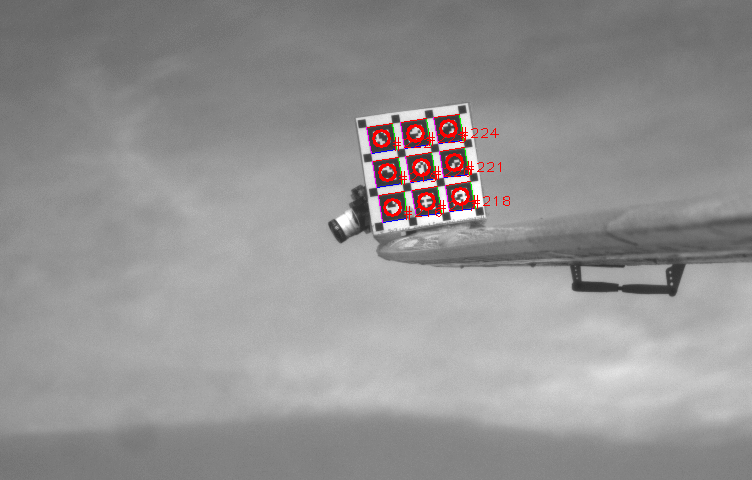}
    \caption{In-air}
  \end{subfigure}
  \caption{Side view from camera rigidly mounted inside the fuselage, looking onto the right wing tip camera.}
  \label{fig:tag}
\end{figure}

The probabilistic wing model used in this paper is represented by a Gaussian mean and covariance for the relative transformation between wing tip (left, right) and center, fuselage IMU.
In order to capture the aerodynamic forces acting on the wing the relative transformation needs to be measured \emph{in-air} as shown by Fig. \ref{fig:tag}:
The pictures are recorded by the right side camera mounted rigidly inside the fuselage looking onto the wing towards the right wing camera (with attached April tag) and illustrate the aerodynamic lift force acting on the wing.

Our proposed calibration process consists of estimating
\begin{enumerate}
\item the rigid transformation $\mathbf{T}^{C_j}_{\text{tag}_j}$ (in the lab under static conditions) and 
\item the time-varying transformation $\mathbf{T}^{C_{\text{side},j}}_{\text{tag}_j}$ (using data recorded during a flight).
\end{enumerate}

\begin{figure}[htb]
\includegraphics[width=\linewidth, trim=0 30 0 0,clip=true]{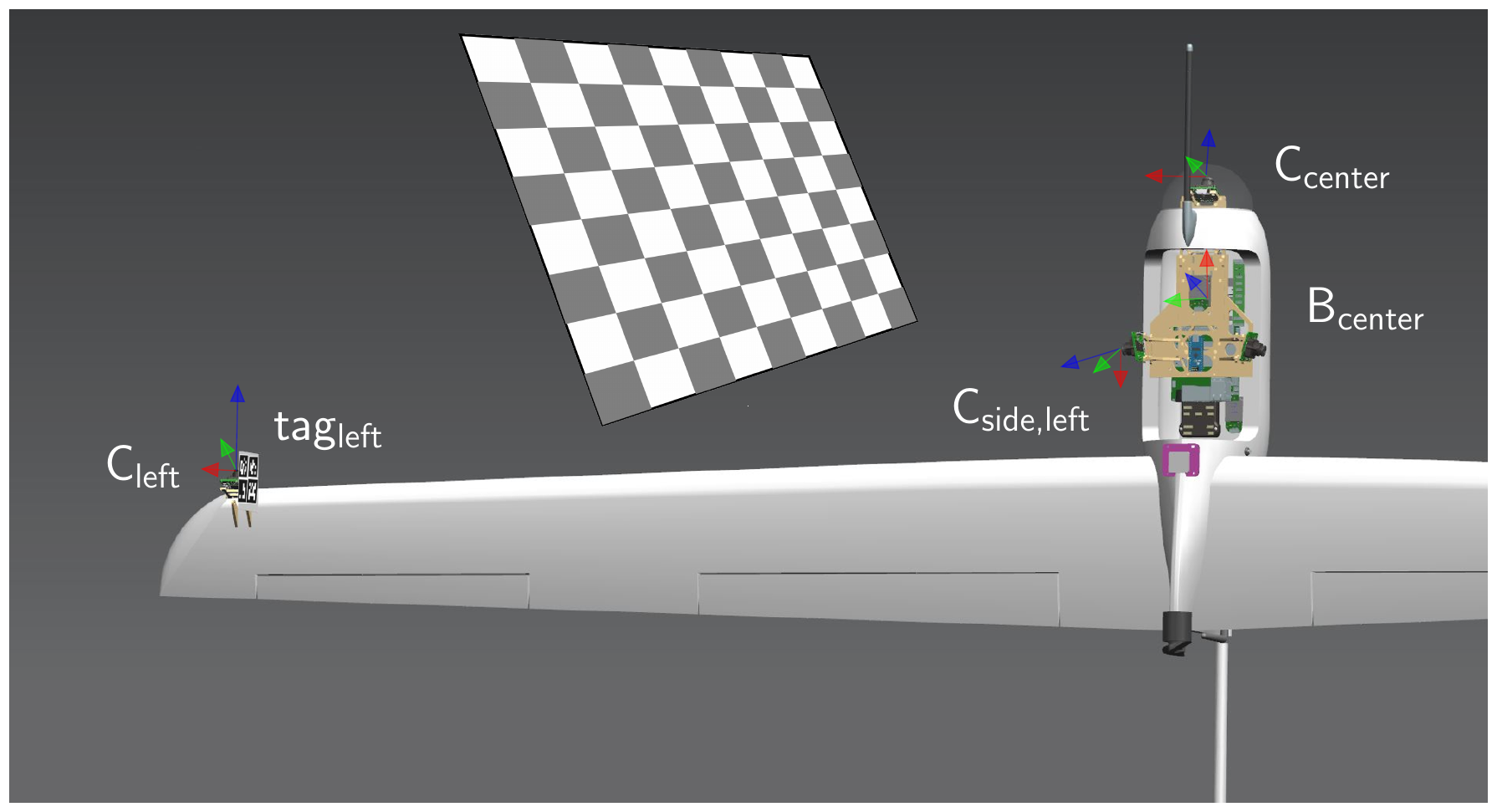}
\caption{Coordinate frames involved in the calibration procedure to obtain a probabilistic wing model (left side).}
\label{fig:calib}
\end{figure}

Firstly, we describe how to obtain the rigid transformation $\mathbf{T}^{C_j}_{\text{tag}_j}$ between the wing camera and the wing tag:
The experiment setup and involved coordinate frames are shown in Fig. \ref{fig:calib}.
The rigid relative transformation between the wing camera and wing tag is given by
\begin{align}
\mathbf{T}^{C_j}_{\text{tag}_j} = 
\mathbf{T}^{C_j}_{C_c}
\mathbf{T}^{C_c}_{C_{\text{side,j}}}
 \mathbf{T}^{C_{\text{side,j}}}_{\text{tag}_j}
\end{align}
where the rigid transformations $\mathbf{T}^{C_j}_{C_c}$  and ${\mathbf{T}^{C_c}_{C_{\text{side,j}}}=\mathbf{T}^{C_c}_{B_c} \mathbf{T}^{B_c}_{C_{\text{side,j}}}}$ are calibrated with \emph{Kalibr} \cite{Rehder2016}.
Note that $\mathbf{T}^{C_j}_{\text{tag}_j}$ could also be obtained by solving the hand-eye problem (e.g. with \cite{Furrer2017}) but requires access to an accurate motion capture system (e.g. \emph{Vicon}).
Instead, our procedure relies only on the synchronized images and inertial measurements from the VI-sensor.

Secondly, during the flight, the April tags on the wing cameras are detected by the left respectively right side camera.
Based on the April tag detection, the relative transformation $\mathbf{T}^{C_{\text{side,j}}}_{\text{tag}_j}$ is obtained from an absolute pose estimator (PNP).
That is, for every frame, the camera pose of the wing camera with respect to the center camera can be computed as
\begin{align}
\mathbf{T}^{C_j}_{C_c}
= \mathbf{T}^{C_j}_{\text{tag}_j} 
\mathbf{T}^{\text{tag}_j}_{C_{\text{side,j}}}
\mathbf{T}^{C_{\text{side,j}}}_{C_c}. \label{eq:wing}
\end{align}
Note that the side cameras and April tags need to be mounted only once for every UAV type in order to establish the wing model.
%
\section{PLATFORM AND HARDWARE}
Fig. \ref{fig:overview} gives an overview of the employed UAV platform \emph{Techpod} and of the VI-sensor system integration.
Due to specific hardware requirements, such as the long baseline between the wing and center camera, commercially available VI-sensors \cite{Nikolic2014} could not be used and a customized VI-sensor based on the \emph{Atmega328P} microcontroller unit (MCU) was developed:
The Camera-IMU rig for the left and right wing each consist of a 
\SI{0.3}{MP} \emph{Aptina MT9V034} camera and an IMU of type \emph{MPU6050}, rigidly mounted behind the camera.
The IMU measurements are triggered via interrupts and transferred to the MCU.
Due to the required length of the data lines, three i$^2$C repeaters are required: one next to the MCU and two next to the Camera-IMU rigs on the wing just before the signal enters the \emph{MPU6050}.
The spatially closer and more accurate center IMU (\emph{ADIS16448}) is connected to the MCU via SPI.
The three IMUs and the five cameras of same type (three in normal mode, two additional for the wing calibration) are time-synchronized via hardware lines and run at \SI{100}{Hz} and up to \SI{20}{Hz}
respectively.
%
The rigid, non-time-varying transformations $\mathbf{T}^{B_\text{px4}}_{B_\text{center}}$  and $\mathbf{T}^{B_j}_{C_j}$ with $j\in\{\text{left},\text{right},\text{center}\}$ are calibrated offline with \emph{Kalibr} \cite{Rehder2016}, where $B_\text{px4}$ refers to the IMU (\emph{ADIS16448}) used by the \emph{pixhawk} Autopilot.
An \emph{UP Squared} with \emph{Intel Atom} CPU at $\unit[1.6]{GHz}$ was used as on-board computer (OBC).
%
%
%

%
\section{EXPERIMENTS}
\subsection{Inertial measurements}
In particular for the inertial measurement units mounted on the wing tips strong vibrations are to be expected.
To quantify this effect, we compare \emph{measured} to \emph{expected} IMU readings.
The camera poses $\mathbf{T}^W_{C_j}$ are optimized offline (using GPS and vision measurements only) and transformed to the IMU poses $\mathbf{T}^W_{B_j}$ via the rigid transformation $\mathbf{T}^{C_j}_{B_j}$.
Using our implementation of \cite{Sittel2013} the expected linear accelerations and angular velocities are computed.
\begin{figure}[htb]
        \begin{subfigure}[b]{0.49\linewidth}
            \includegraphics[width=1\linewidth]{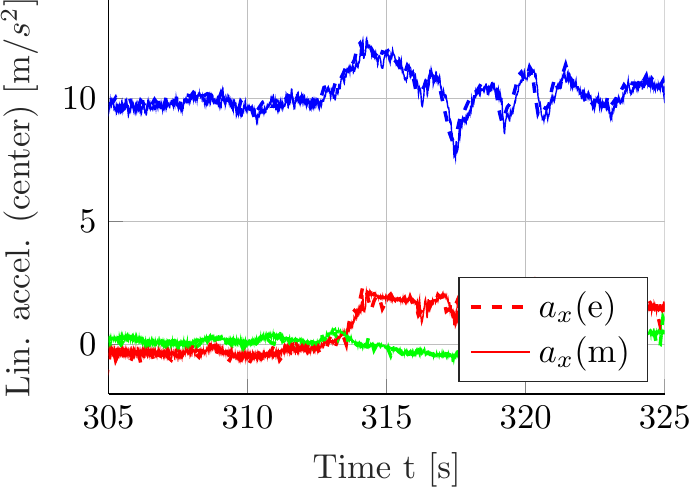}
            \caption[]{}
            \label{fig:inertial_a}
        \end{subfigure}
        \begin{subfigure}[b]{0.49\linewidth}  
            \includegraphics[width=1\linewidth]{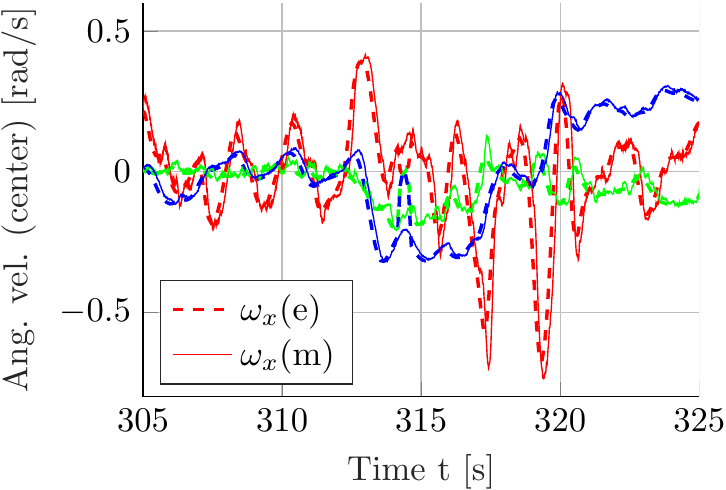}
            \caption[]{}
                        \label{fig:inertial_b}
        \end{subfigure}
        \begin{subfigure}[b]{0.49\linewidth}   
            \includegraphics[width=1\linewidth]{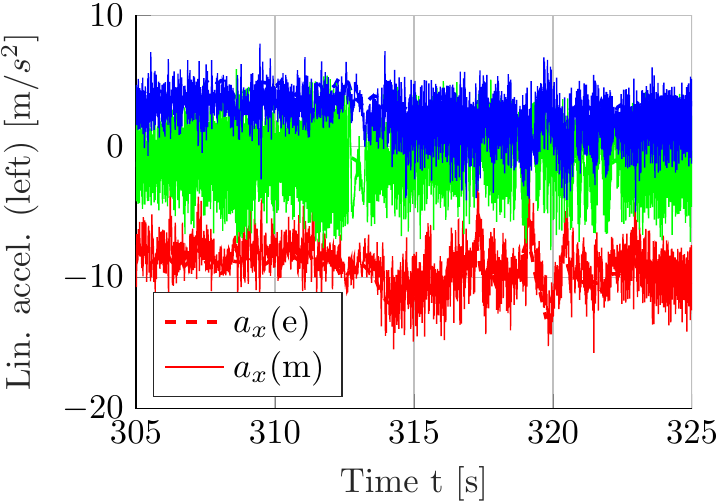}
            \caption[]{}
                        \label{fig:inertial_c}
        \end{subfigure}
        \begin{subfigure}[b]{0.49\linewidth}   
            \includegraphics[width=1\linewidth]{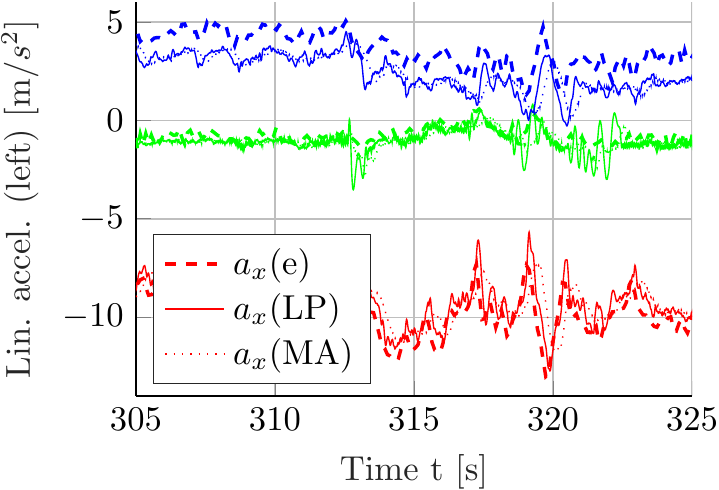} 
            \caption[]{} 
                        \label{fig:inertial_d}
        \end{subfigure}
        \caption[]{Expected (e), measured (m) and filtered (low-pass: LP, moving average: MA) inertial measurements, recorded by the center and left wing tip IMU. The color coding is x (red), y (green), z (blue) throughout the paper.}
        \label{fig:inertial}
\end{figure}
The expected and measured (raw, not de-biased) inertial measurements for the center fuselage camera are shown in Fig. \ref{fig:inertial_a}, \ref{fig:inertial_b}.
The IMU \emph{ADIS16448} is used with a measurement sensitivity of $\pm 500\text{\textdegree}/s$, FIR filter, and four filter taps.
Fig. \ref{fig:inertial_c} shows the measured (raw, not de-biased) accelerometer readings for the left wing IMU using \emph{MPU6050}'s on-board digital low-pass filter (DLPF) with a cut-off frequency $f_c=\SI{94}{\Hz}$.
Fig. \ref{fig:inertial_d} plots the solution from a low-pass filter with $f_c=\SI{3}{\Hz}$ and moving-average filter with window size of $50$ on the right. 
Based on these results, we conclude to use \emph{MPU6050}'s on-board DLPF at $f_c=\SI{5}{\Hz}$ (minimal available cut-off frequency) and compensate for the introduced constant, known delays.
%
\subsection{Probabilistic wing model}
\begin{figure}[htb]
\begin{subfigure}[b]{0.49\linewidth}
\includegraphics[width=1\linewidth]{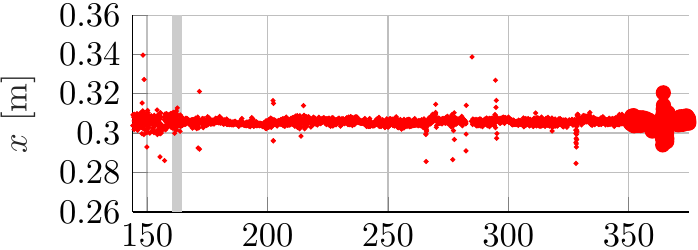}
\label{fig:wing_tx}
\end{subfigure}
\begin{subfigure}[b]{0.49\linewidth}
\includegraphics[width=1\linewidth]{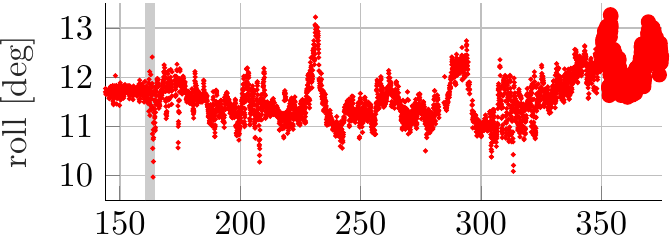}
\label{fig:wing_roll}
\end{subfigure}
\begin{subfigure}[b]{0.49\linewidth}
\includegraphics[width=1\linewidth]{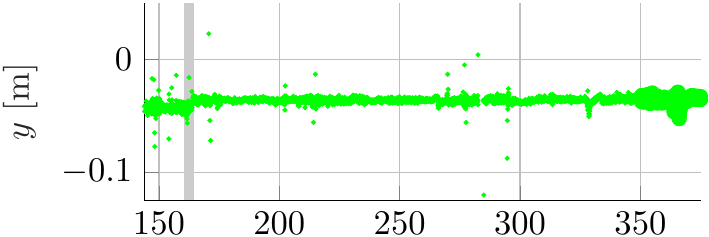}
\label{fig:wing_ty}
\end{subfigure}
\begin{subfigure}[b]{0.49\linewidth}
\includegraphics[width=1\linewidth]{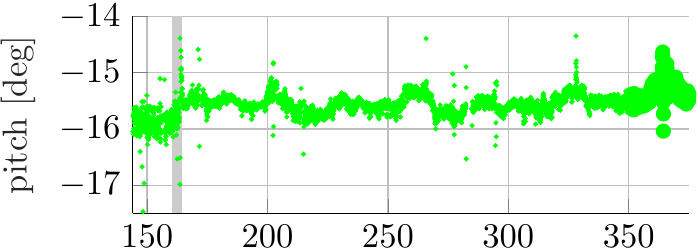}
\label{fig:wing_pitch}
\end{subfigure}
\begin{subfigure}[b]{0.49\linewidth}
\includegraphics[width=1\linewidth]{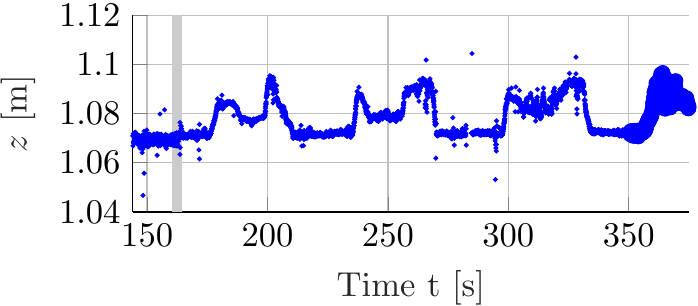}
\label{fig:ting_tz}
\end{subfigure}
\begin{subfigure}[b]{0.49\linewidth}
\includegraphics[width=1\linewidth]{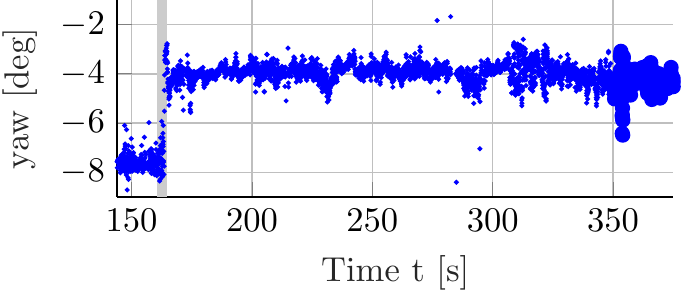}
\label{fig:wing_yaw}
\end{subfigure}
\caption{Observations of the April Tag attached to the right wing camera, as seen from the right side camera. The grey vertical line marks the take-off (landing is not shown).}
\label{fig:wing}
\end{figure}

Fig. \ref{fig:wing} plots the measured relative transformation between the April tag attached to the right wing camera and the right side camera $\mathbf{T}^{\text{tag}_\text{right}}_{C_\text{side,right}}$.
The experiment begins on the ground with take-off approximately at the $\SI{162.5}{\s}$ mark.
The relative translation and orientation vary within $\SI{5}{\centi\metre}$ and $\SI{6}{\degree}$ respectively during the duration of the experiment (including on-ground detections).
The take-off is particularly well observable in the rotation around the $z$-axis.
The observed transformation $\mathbf{T}^{\text{tag}_j}_{C_\text{side,j}}$ is then transformed to $\mathbf{T}^{B_j}_{B_c}$ via the rigid transformation chain given in Eq. \ref{eq:wing} which is stored in the EKF in form of a mean and variance for the left-center and center-right stereo pair.
\subsection{Vision update}
\begin{figure}[b]
  \begin{subfigure}[b]{0.49\linewidth}
    \includegraphics[width=\textwidth]{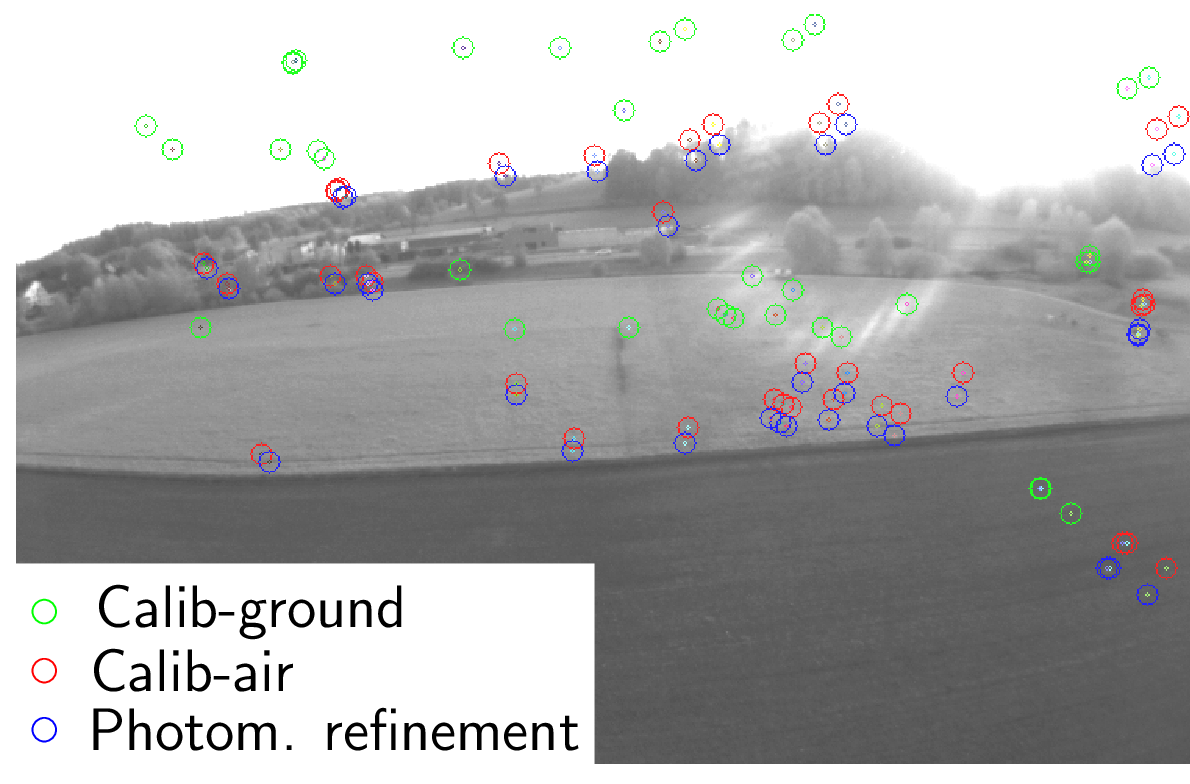}
    \caption{Cam left}
  \end{subfigure}
  \hfill
  \begin{subfigure}[b]{0.49\linewidth}
    \includegraphics[width=\textwidth]{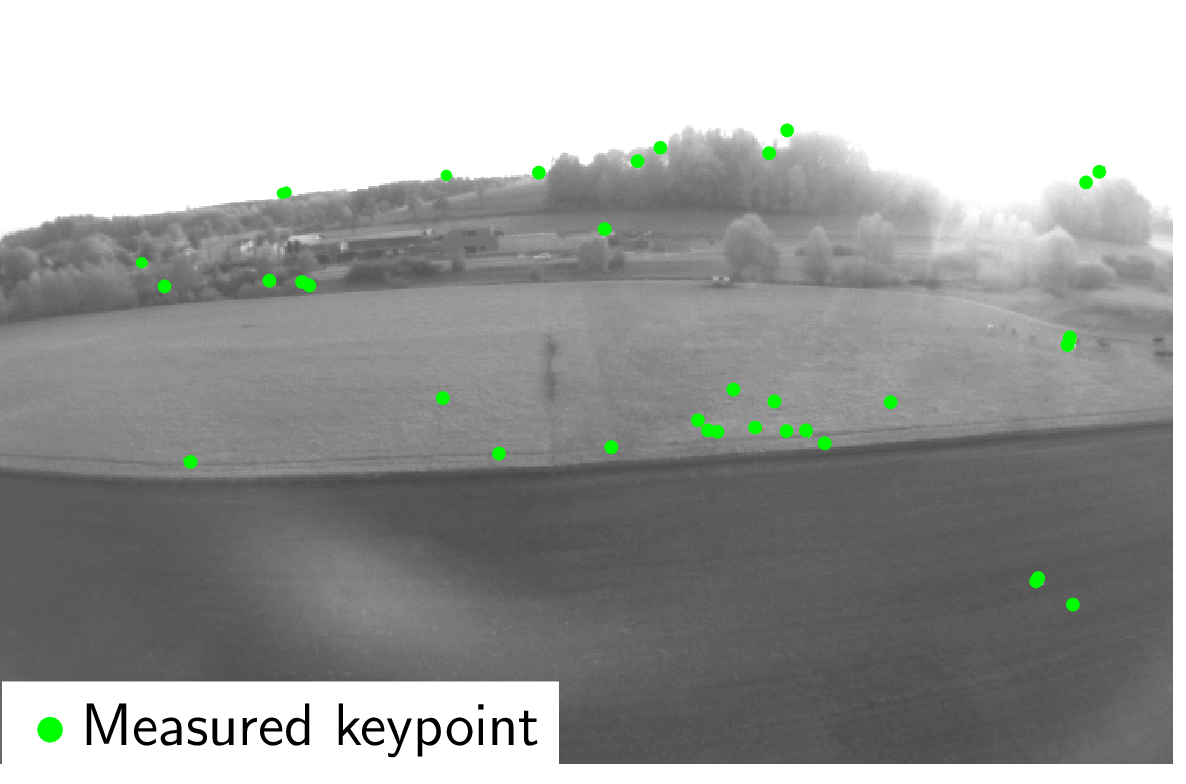}
        \caption{Cam center}
  \end{subfigure}  
\caption{Wing model prior and photometric refinement step for the alignment of the wing and center camera image.}
\label{fig:vision}
\end{figure}

Fig. \ref{fig:vision} visualizes the quality difference between the relative camera-to-camera transformation priors, as well as the photometric alignment of the left to the center camera.
By \emph{Calib-ground} we denote the transformation estimate $\mathbf{T}^{C_l}_{C_c}$ obtained using \emph{Kalibr} in a static setup and with no load applied to the wings.
In contrast, \emph{Calib-air} represents the mean of our wing model (cf. Sec. \ref{secC}).
Features tracked in the center camera are projected into the left camera according to Eq. \ref{eq:depth}:
The features re-projected with the rigid relative transformation $\mathbf{T}^{C_l}_{C_c}$ obtained from \emph{Calib-ground} are shown in green, the features re-projected using the mean of the wing model (\emph{Calib-air}) are shown in red.
%
%
One can observe that the features obtained from \emph{Calib-ground} are far off from the correct feature position, while \emph{Calib-air} returns a relative good initial position in particular during level flight.
The feature positions refined by the photometric sparse image alignment are marked in blue (Eq. \ref{eq:sparse_img_align}).
The photometric formulation shows a robust performance even under challenging conditions such as the lense flare.
\subsection{Depth map}
\begin{figure}[htb]
\captionsetup[subfigure]{justification=centering}
  \begin{subfigure}[b]{0.24\linewidth}
    \frame{\includegraphics[width=\textwidth]{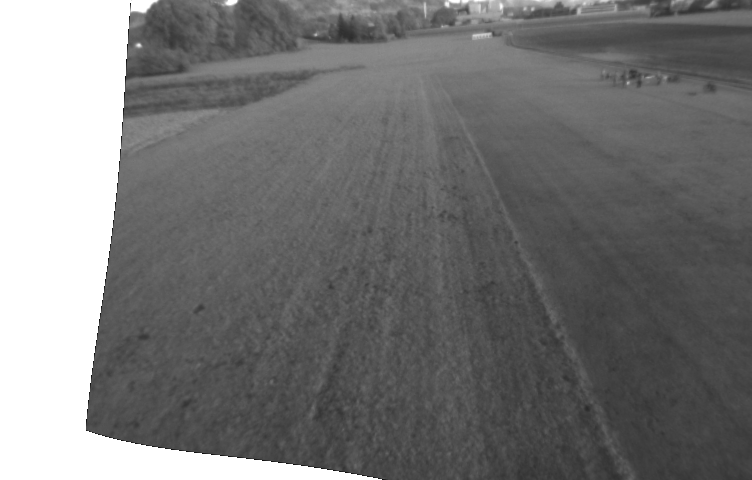}}
  \end{subfigure}
  \hfill
  \begin{subfigure}[b]{0.24\linewidth}
    \frame{\includegraphics[width=\textwidth]{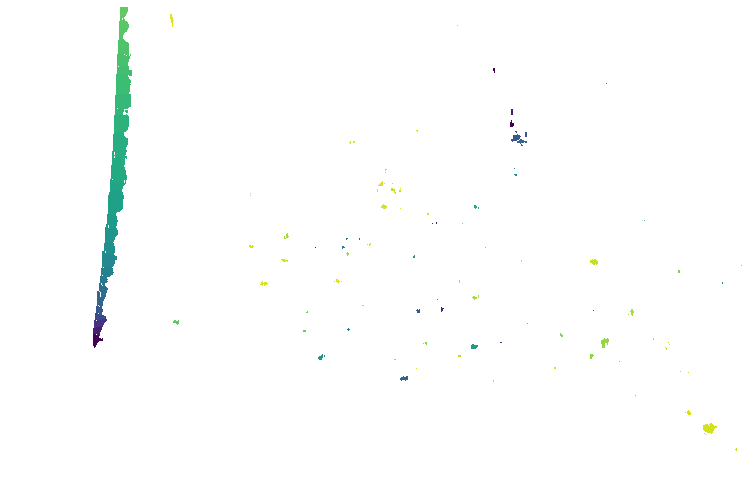}}
  \end{subfigure}
    \begin{subfigure}[b]{0.24\linewidth}
\frame{\includegraphics[width=\textwidth]{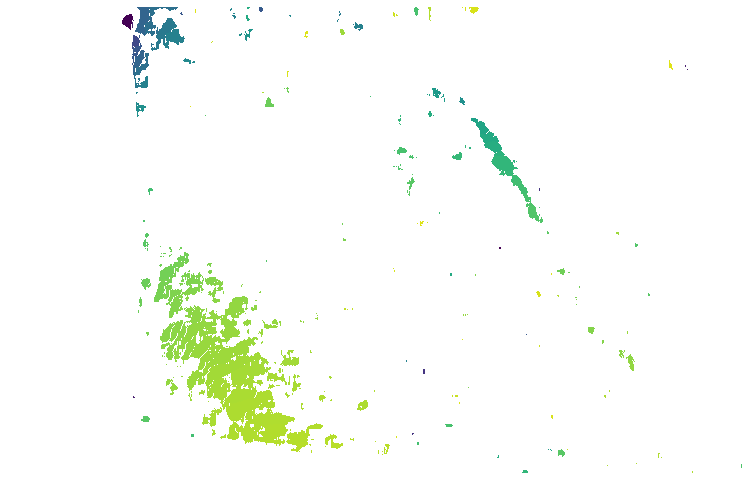}}
  \end{subfigure}
    \begin{subfigure}[b]{0.24\linewidth}
    \frame{\includegraphics[width=\textwidth]{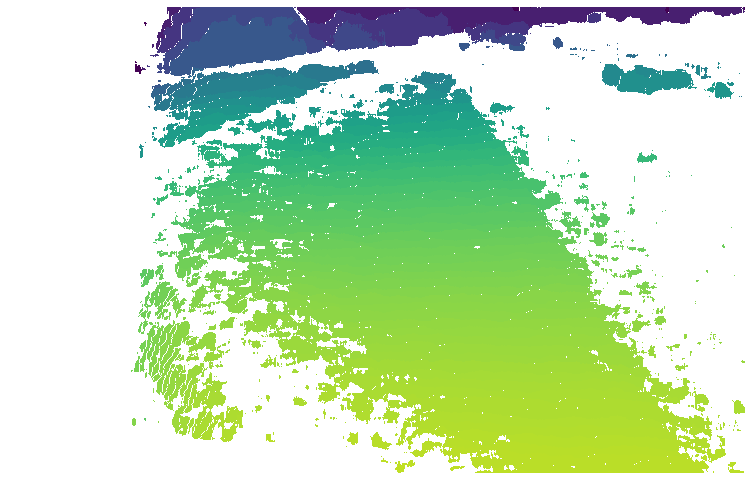}}
  \end{subfigure}
  \\ \vspace{5pt}
   \begin{subfigure}[b]{0.24\linewidth}
    \frame{\includegraphics[width=\textwidth]{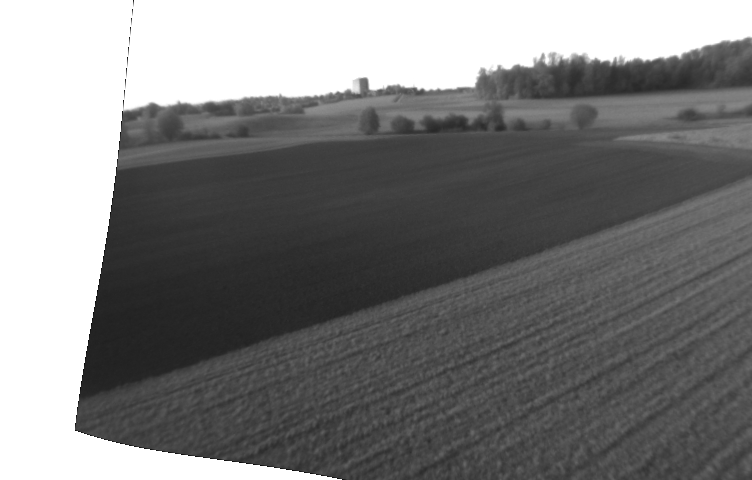}}
  \end{subfigure}
  \hfill
  \begin{subfigure}[b]{0.24\linewidth}
    \frame{\includegraphics[width=\textwidth]{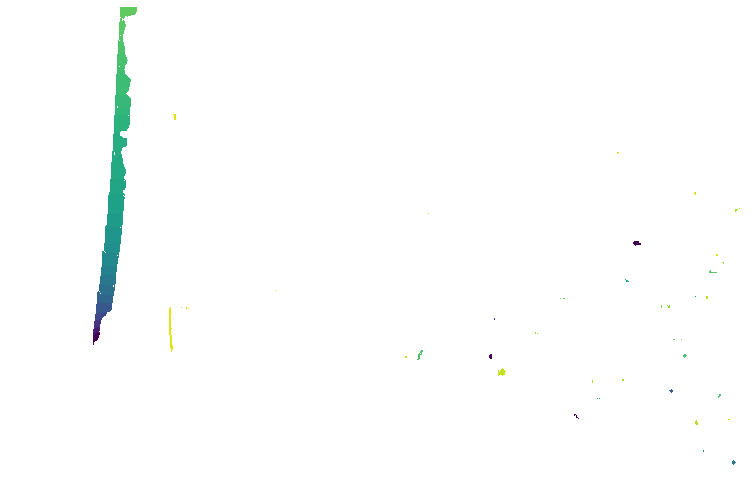}}
  \end{subfigure}
    \begin{subfigure}[b]{0.24\linewidth}
\frame{\includegraphics[width=\textwidth]{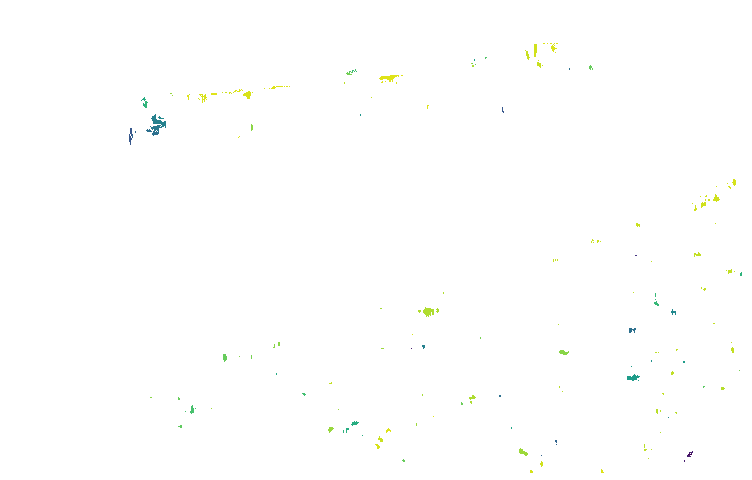}}
  \end{subfigure}
    \begin{subfigure}[b]{0.24\linewidth}
    \frame{\includegraphics[width=\textwidth]{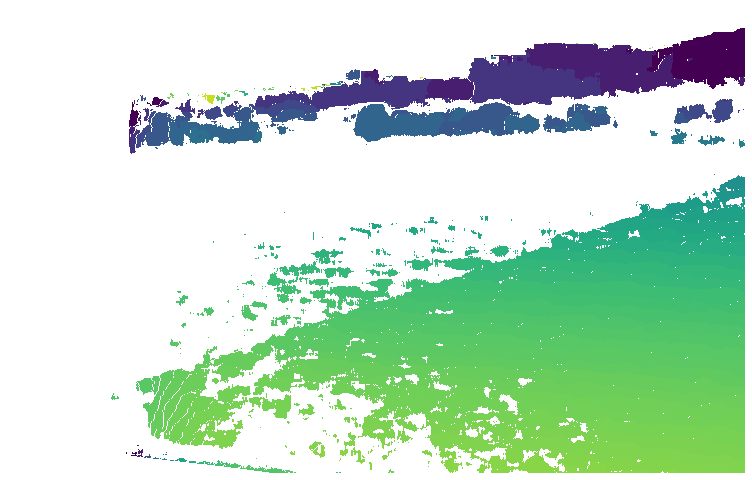}}
  \end{subfigure}
  \\ \vspace{5pt}
  \begin{subfigure}[b]{0.24\linewidth}
    \frame{\includegraphics[width=\textwidth]{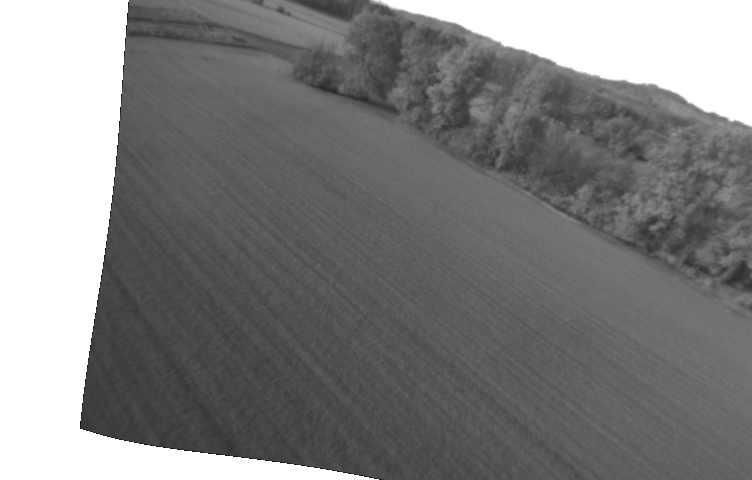}}
        \caption{Rectified center camera}
  \end{subfigure}
  \hfill
  \begin{subfigure}[b]{0.24\linewidth}
    \frame{\includegraphics[width=\textwidth]{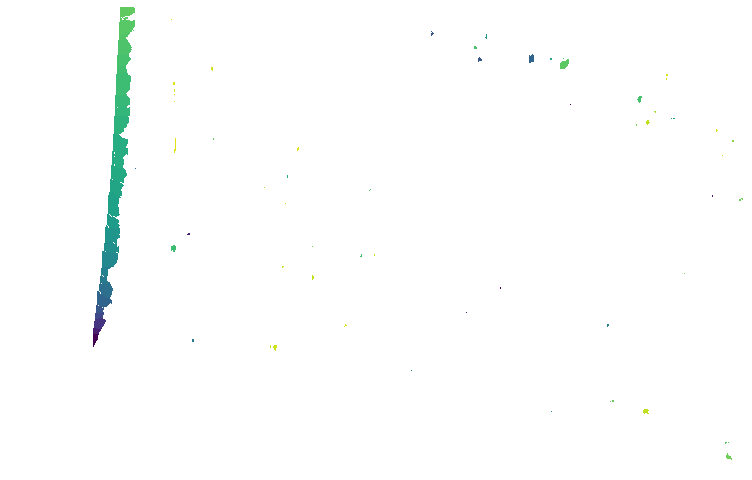}}
        \caption{\\ Calib-ground}
  \end{subfigure}
    \begin{subfigure}[b]{0.24\linewidth}
\frame{\includegraphics[width=\textwidth]{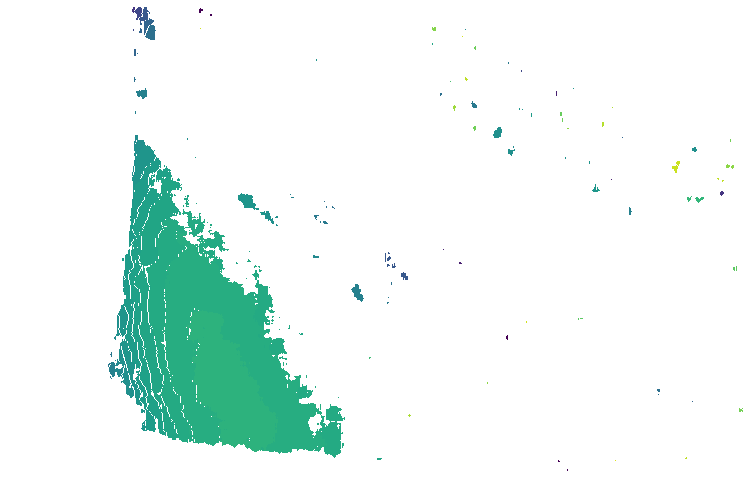}}
\centering
        \caption{\\ Calib-air}
  \end{subfigure}
    \begin{subfigure}[b]{0.24\linewidth}
    \frame{\includegraphics[width=\textwidth]{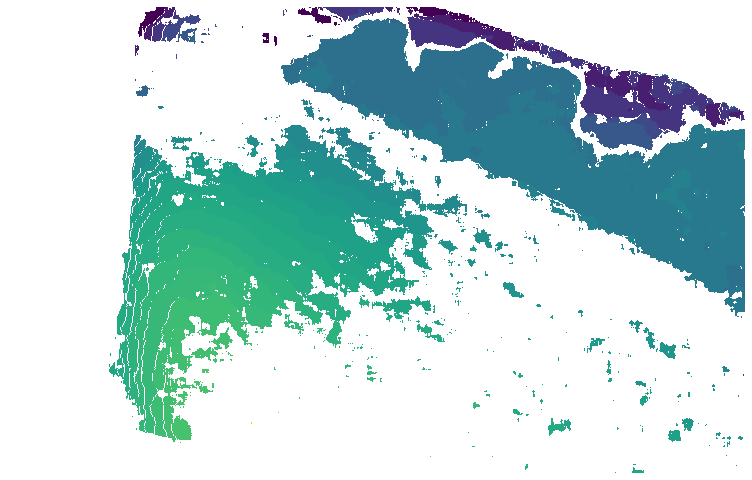}}
        \caption{\\ Proposed}
  \end{subfigure}
  \caption{Single shot depth maps obtained from the left and center image, expressed in the rectified center camera.}
\label{fig:depth_map}
\end{figure}

Fig. \ref{fig:depth_map} depicts the depth maps for the left-center stereo pair, computed using efficient block-matching (BM) \cite{bradski2000opencv}, with $\mathbf{T}^{C_l}_{C_c}$ from b) \emph{Calib-ground}, c) \emph{Calib-air}, and d) our proposed framework.
Areas with low texture variation can potentially be filled in by using different stereo pairs and by observing the area from different viewpoints.

\subsection{Runtime}
The runtime results of the main modules are given in Table~ \ref{table:runtime}, measured on an \emph{Intel i7-4800MQ} CPU at $\unit[2.70]{GHz}$ for comparison. 
\begin{table}[h!]
\scriptsize
\begin{tabular}{|l|l|l|}\hline
\rowcolor{black!30} Module & mean [ms] & Remark \\ \hline
\rowcolor{black!0} Relative IMU integration (0-order) & $0.03$  & Frame to frame  \\\hline 
\rowcolor{black!10} Photometric refinement & $3.2$   & Per frame  \\\hline 
\rowcolor{black!0} Rectification, block-matching & $38.1$  & Per stereo pair  \\\hline 
\end{tabular}
\caption{Runtime results}
\label{table:runtime}
\end{table}
The combination of relative IMU integration, good relative pose prior for the sparse photometric image alignment, and the employed EKF formulation makes the framework extremely efficient.
With the tested frame rate of $\SI{10}{Hz}$ the image stream can be processed with some margin.
Furthermore, the modular formulation makes it possible to run several visual-inertial stereo pairs in different threads.
At this point, the employed rectification and block matching algorithm appears to become the first bottleneck if the image rate is to be set to a higher value.
%
%
%
%
\section{CONCLUSIONS AND FUTURE WORK}
In this paper, we further developed the idea of using visual-inertial sensor systems on-board a UAV for improved environmental awareness and demonstrated the effectiveness in real-world experiments.
We investigated the challenges encountered in the three modules of the framework:
The problem of measuring strong high-frequency vibrations on the wing tip IMUs was solved in software in form of a low-pass filter.
%
%
In future work, we aim at a solution that minimizes the vibrations already on the hardware level.
Since the relative transformation is estimated, a flexible mount or damping material could be used to isolate the visual-inertial system from the motor and wind gust induced vibrations encountered on the wing.

Furthermore, we described our procedure to obtain a probabilistic wing model, formulated as a Gaussian prior, and showed the discrepancy to the on-ground calibration.
In future work, a more sophisticated wing model could be identified.
For instance, the wing could be modeled as a  cantilever beam with additional inputs such as air speed.
As the deformation of the wing influences the lift distribution our model and accelerometer readings could also be incorporated into the on-board controller, e.g. for wind gust rejection.

Finally, in contrast to \cite{Hinzmann2018}, we used a photometric sparse image alignment formulation to compute the relative transformation between wing and center camera.
This approach has shown to be a robust way to transfer and track features from one to another camera under challenging conditions.
In a next step, we intend to test our framework on a platform with stronger wing flapping behavior such as (manned) gliders.
%



%

\section*{ACKNOWLEDGMENT}
We wish to thank: T. Stastny (experiments and discussion), T. Steiger and Y. Allenspach (safety pilots), A. Ruckli (robustification of April tag detection), T. Mantel and A. Melzer (help with VI-Sensor).
%
%
%
\bibliographystyle{ieeetr}
\bibliography{lib.bib}
\end{document}